\pdfoutput=1
\documentclass[11pt]{article}
\usepackage[final]{coling}

\usepackage{times}
\usepackage{latexsym}
\usepackage{adjustbox}
\usepackage{amssymb}
\usepackage{pifont}
\newcommand{\cmark}{\ding{51}}%
\newcommand{\xmark}{\ding{55}}%

\usepackage[T1]{fontenc}
\usepackage[utf8]{inputenc}
\usepackage{microtype}
\usepackage{inconsolata}
\usepackage{microtype}
\usepackage{inconsolata}
\usepackage{graphicx}
\usepackage{url}
\usepackage{times}
\usepackage{latexsym}
\usepackage{longtable}
\usepackage{booktabs}
\usepackage{enumitem}
\usepackage{multirow}
\usepackage{tikz}
\usepackage{amsmath}
\usepackage[ruled, noend]{algorithm2e} 
\usepackage{tabularray}
\usepackage{centernot}
\usepackage{placeins}
\usepackage{amsfonts}
\usepackage[table]{colortbl}
\usepackage{tabularx}
\usepackage{lscape}
\usepackage{array}
\usepackage{cleveref}
\usepackage{tcolorbox}
\usepackage{hyperref}
\usepackage{xcolor}
\usepackage{amsfonts}
\usepackage{subcaption}
\usepackage{ulem}
\newcommand{\method}{\textsc{Con-ReCall}}
\newcommand{\recall}{\textbf{\textit{ReCall}}}

\newcommand{\loss}{\citep{loss_mia}}
\newcommand{\reference}{\citep{ref_mia}}
\newcommand{\zlib}{\citep{zlib_mia}}
\newcommand{\neighbor}{\citep{neighbor_mia}}
\newcommand{\Mink}{\citep{mink}}
\newcommand{\MinkPlusPlus}{\citep{minkpp}}
\newcommand{\recallcite}{\citep{recall_mia}}

\title{\method: Detecting Pre-training Data in LLMs via Contrastive Decoding}

\author{Cheng Wang$^\dagger$, Yiwei Wang$^{||}$ , Bryan Hooi$^\dagger$,  Yujun Cai$^\ddagger$, \textbf{Nanyun Peng}$^\mathsection$, \textbf{Kai-Wei Chang}$^\mathsection$
\\
$^\dagger$ National University of Singapore  
$^{||}$  University of California, Merced\\
$^\mathsection$ University of California, Los Angeles \quad
$^\ddagger$ University of Queensland \\
\texttt{wcheng@comp.nus.edu.sg} 
\\
}

\begin{document}

\maketitle

\begin{abstract}

The training data in large language models is key to their success, but it also presents privacy and security risks, as it may contain sensitive information. Detecting pre-training data is crucial for mitigating these concerns. 
Existing methods typically analyze target text in isolation or solely with non-member contexts, overlooking potential insights from simultaneously considering both member and non-member contexts. 
While previous work suggested that member contexts provide little information due to the minor distributional shift they induce, our analysis reveals that these subtle shifts can be effectively leveraged when contrasted with non-member contexts.
In this paper, we propose \method{}, a novel approach that leverages the asymmetric distributional shifts induced by member and non-member contexts through contrastive decoding, amplifying subtle differences to enhance membership inference. Extensive empirical evaluations demonstrate that \method{} achieves state-of-the-art performance on the WikiMIA benchmark and is robust against various text manipulation techniques.~\footnote{Our code is available at the following repo: \href{https://github.com/WangCheng0116/CON-RECALL}{{https://github.com/WangCheng0116/CON-RECALL}}
}
\end{abstract}

\section{Introduction}

Large Language Models (LLMs)~\citep{gpt4, llama2} have revolutionized natural language processing by achieving remarkable performance across a wide range of language tasks.
These models owe their success to extensive training datasets, often encompassing trillions of tokens.
However, the sheer volume of these datasets makes it practically infeasible to meticulously filter out all inappropriate data points.
Consequently, LLMs may unintentionally memorize sensitive information, raising significant privacy and security concerns.
This memorization can include test data from benchmarks \citep{sainz-etal-2023-nlp, oren2023provingtestsetcontamination}, copyrighted materials \citep{meeus2023didneuronsreadbook, duarte2024decopdetectingcopyrightedcontent, chang-etal-2023-speak}, and personally identifiable information \citep{mozes2023usellmsillicitpurposes, tang2024privacypreservingincontextlearningdifferentially}, leading to practical issues such as skewed evaluation results, potential legal ramifications, and severe privacy breaches.
Therefore, developing effective techniques to detect unintended memorization in LLMs is crucial.

\begin{figure}[t]
\includegraphics[width=0.48\textwidth]{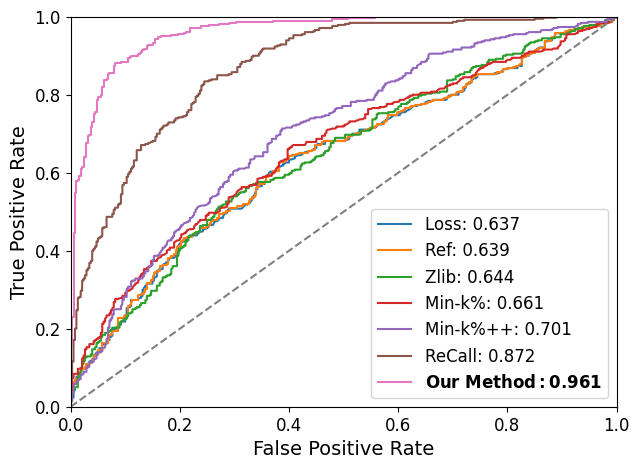}
\caption{\textbf{AUC performance on WikiMIA-32 dataset.} Our \method{} significantly outperforms the current state-of-the-art baselines.
}
\label{fig:roc}
\vspace{-1em}
\end{figure}

Existing methods for detecting pre-training data~\citep{loss_mia, minkpp, recall_mia} typically analyze target text either in isolation or alongside with non-member contexts, while commonly neglecting member contexts. 
This omission is based on the belief that member contexts induce only minor distributional shifts, offering limited additional value~\citep{recall_mia}.

\begin{table*}[ht]
\centering
\begin{adjustbox}{width=\linewidth}
\renewcommand{\arraystretch}{1.1}
\begin{tabular}{lcc}
\toprule
\textbf{Method} & \textbf{Formula} & \textbf{Reference Based} \\
\midrule
\textbf{\textit{Loss}}~\loss & $\mathcal{L}(x, \mathcal{M})$ & \xmark \\
\textbf{\textit{Ref}}~\reference & $\mathcal{L}(x, \mathcal{M}) - \mathcal{L}(x, \mathcal{M}_{ref})$ & \cmark \\
\textbf{\textit{Zlib}}~\zlib & $\frac{\mathcal{L}(x, \mathcal{M})}{zlib(x)}$ & \xmark \\
\textbf{\textit{Neighborhood Attack}}~\neighbor & $\mathcal{L}(x; \mathcal{M}) - \frac{1}{n}\sum_{i=1}^n \mathcal{L}(\tilde{x}_i; \mathcal{M})$ & \xmark \\
\textbf{\textit{Min-K\%}}~\Mink & $
\frac{1}{\lvert \text{min-}k(x)\rvert} \sum_{x_i \in \text{min-}k(x)} -\log(p(x_i \mid x_1, \ldots, x_{i-1}))$ & \xmark \\
\textbf{\textit{Min-K\%++}}~\MinkPlusPlus & \begin{tabular}{@{}c@{}}
$\text{Min-K\%++}_{\text{token}}(x_t) = \frac{\log p(x_t \mid x_{<t}) - \mu_{x_{<t}}}{\sigma_{x_{<t}}},$ \\ \\
$\text{Min-K\%++}(x) = \frac{1}{\lvert \text{min-k\%} \rvert} \sum_{x_t \in \text{min-k\%}} \text{Min-K\%++}_{\text{token}}(x_t)$
\end{tabular} & \xmark \\
\textbf{\textit{ReCall}}~\recallcite & $\frac{LL(x|P_{\text{non-member}})}{LL(x)}$ & \xmark \\
\bottomrule
\end{tabular}
\end{adjustbox}
\caption{\textbf{Comparison of baseline methods.} This table provides an overview of different membership inference methods, their mathematical formulations, and whether they require a reference model.}
\label{tab:methods_comparison}
\vspace{0.5em}
\end{table*}

However, our analysis reveals that these subtle shifts in member contexts, though often dismissed, hold valuable information that has been underexploited. The central insight of our work is that information derived from member contexts gains significant importance when contrasted with non-member contexts. This observation led to the development of \method{}, a novel approach that harnesses the contrastive power of prefixing target text with both member and non-member contexts. By exploiting the asymmetric distributional shifts induced by these different prefixes, \method{} provides more nuanced and reliable signals for membership inference. This contrastive strategy not only uncovers previously overlooked information but also enhances the accuracy and robustness of pre-training data detection, offering a more comprehensive solution than existing methods.

To demonstrate the effectiveness of \method{},
we conduct extensive empirical evaluations
on the method across a variety of models of different sizes. Our experiments show that \method{} outperforms the current state-of-the-art method by a significant margin, as shown in Figure~\ref{fig:roc}. Notably, \method{} only requires a gray-box access to LLMs, i.e., token probabilities, and does not necessitate a reference model, enhancing its applicability in real-world scenarios.

\begin{figure*}[t]
\centering
\includegraphics[width=0.96\textwidth]{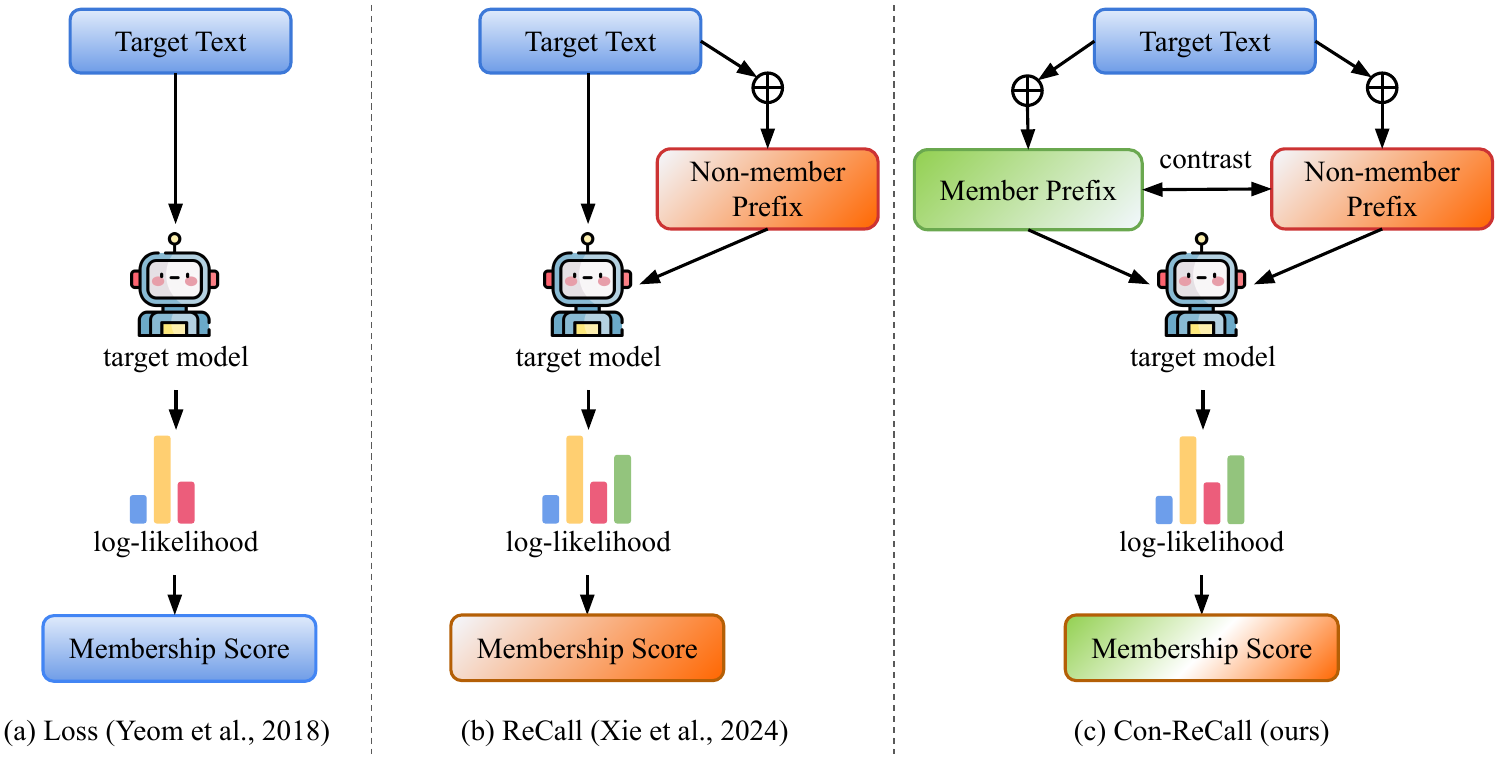}
\caption{\textbf{Overview of three MIA methods}. Our method refines the previous membership score by incorporating contrastive information when prefixing target text with members and non-members. }
\label{fig:framework}
\vspace{-1em}
\end{figure*}

\section{Related Work}

\paragraph{Detecting Pre-training Data in LLMs.}
While membership inference attacks (MIA) have been extensively studied in various domains~\citep{first_mia, carlini2023extractingtrainingdatadiffusion, watson2022importancedifficultycalibrationmembership}, detecting pre-training data in LLMs presents unique challenges. Unlike classical MIA, LLM developers rarely release full training data~\citep{gpt4, llama2}, and single-epoch training on vast datasets makes memorization detection difficult~\citep{carlini2023quantifyingmemorizationneurallanguage, mink}. \citet{mink} pioneered this research with the WikiMIA benchmark and Min-K\% baseline method. \citet{minkpp} improved Min-K\% through token log-probability normalization, while the ReCall method~\recallcite{} currently achieves state-of-the-art performance using relative conditional log-likelihoods. These methods contribute to the broader application of MIA in detecting copyrighted materials, personally identifiable information, and test-set contamination~\citep{meeus2023didneuronsreadbook, mozes2023usellmsillicitpurposes, sainz-etal-2023-nlp}.

\paragraph{Contrastive Decoding.}

Contrastive decoding is primarily a method for text generation. Depending on the elements being contrasted, it serves different purposes. For example, DExperts~\citep{liu-etal-2021-dexperts} use outputs from a model exposed to toxicity to guide the target model away from undesirable outputs. Context-aware decoding~\citep{shi-etal-2024-trusting} contrasts model outputs given a query with and without relevant context. \citet{zhao-etal-2024-enhancing} further enhance context-aware decoding by providing irrelevant context in addition to relevant context. In this paper, we adapt the idea of contrastive decoding to MIA, where the contrast occurs between target data prefixed with member and non-member contexts. 

\section{\method}

\subsection{Problem Formulation}

Consider a model $\mathcal{M}$ trained on dataset $\mathcal{D}$. The objective of a membership inference attack is to ascertain whether a data point $x$ belongs to $\mathcal{D}$ (i.e., $x \in \mathcal{D}$) or not (i.e., $x \notin \mathcal{D}$). Formally, we aim to develop a scoring function $s(x, \mathcal{M}) \rightarrow \mathbb{R}$, where the membership prediction is determined by a threshold $\tau$:
$$
\begin{cases}
  x \in \mathcal{D} & \text{if } s(x, \mathcal{M}) \geq \tau \\
  x \notin \mathcal{D} & \text{if } s(x, \mathcal{M}) < \tau \\
\end{cases}.
$$

\begin{figure*}
    \centering
    \includegraphics[width=0.94\linewidth]{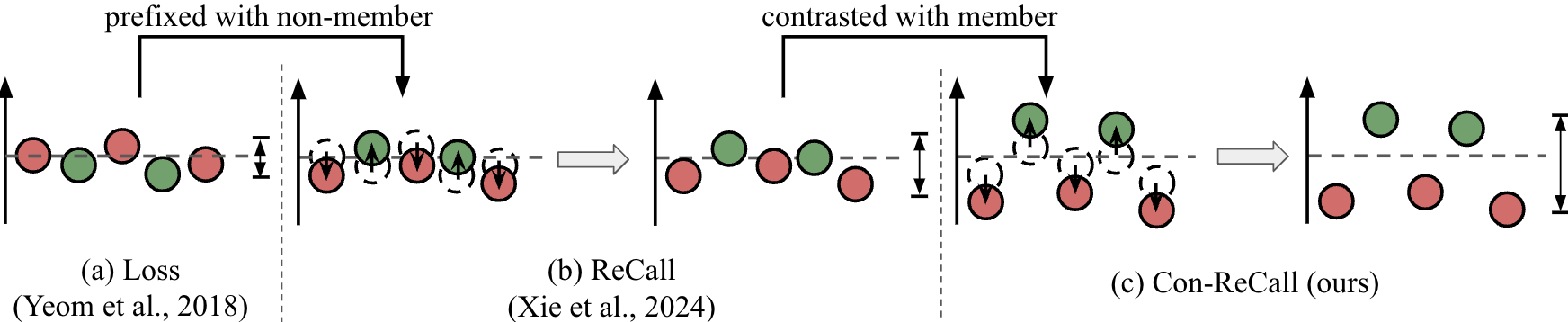}
    \caption{\textbf{Distribution shifts induced by three methods.} (a) Loss directly uses log-likelihoods, resulting in no shift. (b) ReCall examines the shift caused by non-member prefixes. (c) Our \method{} enhances the distinction by contrasting with both member and non-member prefixes.}
    \label{fig:conceptual_illustration}
\end{figure*}

\begin{figure*}[t]
\centering
\includegraphics[width=1\textwidth]{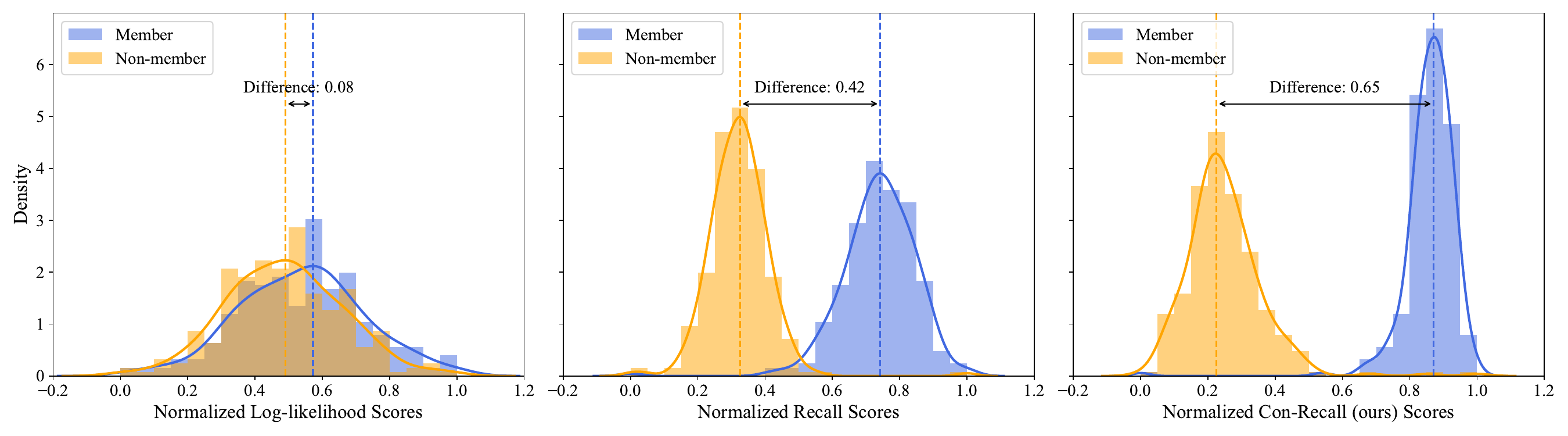}
\caption{\textbf{Visualization of membership score distributions.} Min-max normalized distributions are shown for log-likelihood (left), ReCall (middle), and \method{} (right). \method{} achieves the largest separation between members and non-members.}
\label{fig:distribution}
\vspace{-1em}
\end{figure*}

\subsection{Motivation}

Our key insight is that prefixing target text with contextually similar content increases its log-likelihood, while dissimilar content decreases it. Member prefixes boost log-likelihoods for member data but reduce them for non-member data, with non-member prefixes having the opposite effect. This principle stems from language models' fundamental tendency to generate contextually consistent text. 

To quantify the impact of different prefixes, we use the Wasserstein distance to measure the distributional shifts these prefixes induce. For discrete probability distributions $P$ and $Q$ defined on a finite set $X$, the Wasserstein distance $W$ is given by:

\[
W(P, Q) = \sum_{x\in X} |F_P(x) - F_Q(x)|,
\]
where $F_P$ and $F_Q$ are the cumulative distribution functions of $P$ and $Q$ respectively. To capture the directionality of the shift, we introduce a signed variant of this metric:

\[
W_{\text{signed}}(P, Q) = \text{sign}(\mathbb{E}_Q[X] - \mathbb{E}_P[X]) \cdot W(P, Q).
\]

\begin{figure}[ht]
    \centering
    \includegraphics[width=0.47\textwidth]{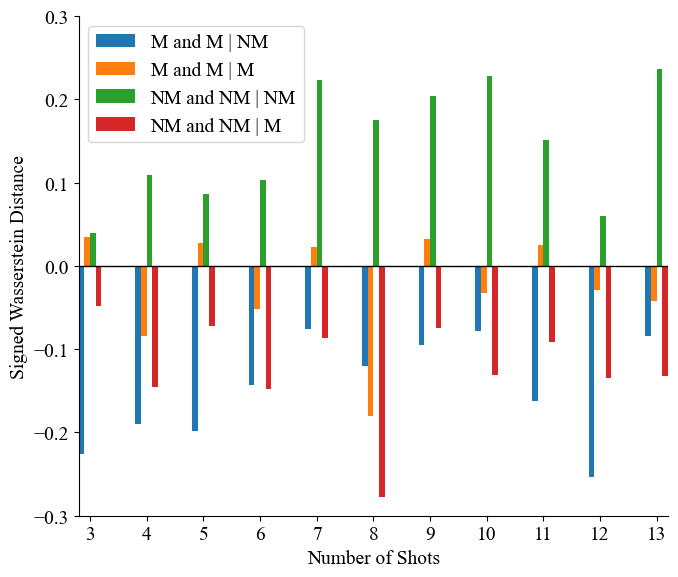}
    \caption{\textbf{Signed Wasserstein distances between original and prefixed distributions across varying shot numbers.} The plot illustrates how the distributional shift, measured by signed Wasserstein distance, changes for member and non-member data when prefixed with different contexts (M: member, NM: non-member).}
    \label{fig:wasserstein_distances}
\end{figure}

Our experiments reveal striking asymmetries in how member and non-member data respond to different prefixes. Figure~\ref{fig:wasserstein_distances} illustrates these asymmetries, showing the signed Wasserstein distances between original and prefixed distributions across varying numbers of shots, where shots refer to the number of non-member data points used in the prefix.

We observe two key phenomena:

\begin{enumerate}
    \item \textbf{Asymmetric Shift Direction}: Member data exhibits minimal shift when prefixed with other member contexts, indicating a degree of distributional stability. However, when prefixed with non-member contexts, it undergoes a significant negative shift. In contrast, non-member data displays a negative shift when prefixed with member contexts and a positive shift with non-member prefixes.
    
    \item \textbf{Asymmetric Shift Intensity}: Non-member data demonstrated heightened sensitivity to contextual modifications, manifesting as larger magnitude shifts in the probability distribution, regardless of the prefix type. Member data, while generally more stable, still exhibited notable sensitivity, particularly to non-member prefixes.
\end{enumerate}

These results corroborate our initial analysis and establish a robust basis for our contrastive approach. The asymmetric shifts in both direction and intensity provide crucial insights for developing a membership inference technique that leverages these distributional differences effectively.

\subsection{Contrastive Decoding with Member and Non-member Prefixes}

Building on the insights from our analysis, we propose \method{}, a method that exploits the contrastive information between member and non-member prefixes to enhance membership inference through contrastive decoding. Our approach is directly motivated by the two key observations from the previous section:

\begin{enumerate}
    \item The asymmetric shift direction suggests that comparing the effects of member and non-member prefixes could provide a strong signal for membership inference.
    \item The asymmetric shift intensity indicates the need for a mechanism to control the relative importance of these effects in the decoding process.
\end{enumerate}

These insights lead us to formulate the membership score $s(x, M)$ for a target text $x$ and model $M$ as follows:

\[
\frac{LL(x|P_{\text{non-member}}) - \gamma \cdot LL(x|P_{\text{member}})}{LL(x)}
,\]
where $LL(\cdot)$ denotes the log-likelihood, $P_{member}$ and $P_{non-member}$ are prefixes composed of member and non-member contexts respectively, and $\gamma$ is a parameter controlling the strength of the contrast.

This formulation provides a robust signal for membership inference by leveraging the distributional differences revealed in our analysis. Figure~\ref{fig:conceptual_illustration} illustrates how our contrastive approach amplifies the distributional differences

Importantly, \method{} requires only gray-box access to the model, utilizing solely token probabilities. This characteristic enhances its practical utility in real-world applications where full model access may not be available, making it a versatile tool for detecting pre-training data in large language models.

\section{Experiments}

In this section, we will evaluate the effectiveness of \method{} across various experimental settings, demonstrating its superior performance compared to existing methods.

\subsection{Setup}

\paragraph{Baselines.}

In our experiment, we evaluate \method{} against seven baseline methods. \textbf{\textit{Loss}}~\loss{} directly uses the loss of the input as the membership score. \textbf{\textit{Ref}}~\reference{} requires another reference model, which is trained on a dataset with a distribution similar to $\mathcal{D}$, to calibrate the loss calculated in the \textbf{\textit{Loss}} method. \textbf{\textit{Zlib}}~\zlib{} instead calibrates the loss by using the input's Zlib entropy. \textbf{\textit{Neighbor}}~\neighbor{} perturbs the input sequence to generate $n$ neighbor data points, and the loss of $x$ is compared with the average loss of the $n$ neighbors. \textbf{\textit{Min-K\%}}~\Mink{} is based on the intuition that a member sequence should have few outlier words with low probability; hence, the top-k\% words having the minimum probability are averaged as the membership score. \textbf{\textit{Min-K\%++}}~\MinkPlusPlus{} is a normalized version of \textbf{\textit{Min-K\%}} with some improvements. \recall{}~\recallcite{} calculates the relative conditional log-likelihood between $x$ and $x$ prefixed with a non-member contexts \( P_{\text{non-member}} \). More details can be found in Table~\ref{tab:methods_comparison}.

\paragraph{Datasets.}

We primarily use WikiMIA~\citep{mink} as our benchmark. WikiMIA consists of texts from Wikipedia, with members and non-members determined using the knowledge cutoff time, meaning that texts released after the knowledge cutoff time of the model are naturally non-members. WikiMIA is divided into three subsets based on text length, denoted as WikiMIA-32, WikiMIA-64, and WikiMIA-128. 

Another more challenging benchmark is MIMIR~\citep{duan2024membership}, which is derived from the Pile~\citep{pile} dataset. The benchmark is constructed using a train-test split, effectively minimizing the temporal shift present in WikiMIA, thereby ensuring a more similar distribution between members and non-members. More details about these two benchmarks are presented in Appendix~\ref{app:stats}.

\paragraph{Models.}

For the WikiMIA benchmark, we use Mamba-1.4B~\citep{gu2024mambalineartimesequencemodeling}, Pythia-6.9B~\citep{biderman2023pythiasuiteanalyzinglarge}, GPT-NeoX-20B~\citep{black-etal-2022-gpt}, and LLaMA-30B~\citep{touvron2023llamaopenefficientfoundation}, consistent with~\citet{recall_mia}. For the MIMIR benchmark, we use models from the Pythia family, specifically 2.8B, 6.9B, and 12B. Since Ref~\citep{ref_mia} requires a reference model, we use the smallest version of the model from that series as the reference model, for example, Pythia-70M for Pythia models, consistent with previous works~\citep{mink, minkpp, recall_mia}. 

\begin{table*}[t!]
\centering
\resizebox{1\textwidth}{!}{%
\begin{tabular}{llcccccccccccc}
\toprule
\multirow{2}{*}{\textbf{Len.}} & \multirow{2}{*}{\textbf{Method}} & \multicolumn{2}{c}{\textbf{Mamba-1.4B}} & \multicolumn{2}{c}{\textbf{Pythia-6.9B}} & \multicolumn{2}{c}{\textbf{NeoX-20B}} & \multicolumn{2}{c}{\textbf{LLaMA-30B}} & \multicolumn{2}{c}{\textbf{Average}} \\
\cmidrule(lr){3-4} \cmidrule(lr){5-6} \cmidrule(lr){7-8} \cmidrule(lr){9-10} \cmidrule(lr){11-12}
 &  & \textbf{AUC} & \textbf{TPR@5\%FPR} & \textbf{AUC} & \textbf{TPR@5\%FPR} & \textbf{AUC} & \textbf{TPR@5\%FPR} & \textbf{AUC} & \textbf{TPR@5\%FPR} & \textbf{AUC} & \textbf{TPR@5\%FPR} \\
\midrule
\multirow{8}{*}{32}
& Loss~\loss & 60.9 & 13.2 & 63.7 & 14.5 & 68.9 & 20.8 & 69.4 & 18.2 & 65.7 & 16.7 \\
& Ref~\reference & 61.2 & 13.4 & 63.9 & 13.7 & 69.1 & 20.3 & 69.9 & 18.7 & 66.0 & 16.5 \\
& Zlib~\zlib & 62.1 & 15.0 & 64.4 & 16.3 & 69.3 & 20.5 & 69.9 & 14.7 & 66.4 & 16.6 \\
& Neighbor~\neighbor & 64.1 & 11.9 & 65.8 & 16.5 & 70.2 & 22.2 & 67.6 & 9.3 & 66.9 & 15.0 \\
& Min-K\%~\Mink & 63.2 & 13.9 & 66.1 & 17.1 & 72.0 & 28.7 & 70.1 & 19.5 & 67.9 & 19.8 \\
& Min-K\%++~\MinkPlusPlus & 66.8 & 12.1 & 70.0 & 13.7 & 75.7 & 17.9 & 84.6 & 27.1 & 74.3 & 17.7 \\
& ReCall~\recallcite & 88.6 & 43.2 & 87.0 & 42.9 & 86.7 & 44.7 & 91.4 & 49.7 & 88.4 & 45.1 \\
\rowcolor{gray!15}& \method{} (ours)  &  \textbf{94.4} & \textbf{68.4} & \textbf{96.0} & \textbf{77.1} & \textbf{95.2} & \textbf{67.6} & \textbf{97.4} & \textbf{87.4} & \textbf{95.8} & \textbf{75.1} \\
\midrule
\multirow{8}{*}{64}
& Loss~\loss & 57.8 & 9.6 & 60.3 & 13.1 & 66.1 & 16.7 & 66.1 & 14.7 & 62.6 & 13.5 \\
& Ref~\reference & 58.1 & 10.0 & 60.4 & 13.5 & 66.3 & 15.5 & 67.0 & 15.5 & 63.0 & 13.6 \\
& Zlib~\zlib & 59.5 & 12.7 & 61.6 & 13.9 & 67.3 & 17.5 & 67.1 & 16.3 & 63.9 & 15.1 \\
& Neighbor~\neighbor & 60.6 & 8.8 & 63.2 & 10.9 & 67.1 & 13.0 & 67.1 & 9.9 & 64.5 & 10.7 \\
& Min-K\%~\Mink & 61.7 & 18.7 & 64.6 & 17.1 & 72.5 & 27.1 & 68.5 & 17.1 & 66.8 & 20.0 \\
& Min-K\%++~\MinkPlusPlus & 66.9 & 13.1 & 71.4 & 15.1 & 76.3 & 23.5 & 85.3 & 34.7 & 75.9 & 21.6 \\
& ReCall~\recallcite & 91.0 & 51.0 & 90.6 & 47.4 & 90.0 & 45.0 & 92.7 & 51.4 & 91.1 & 48.7 \\
\rowcolor{gray!15}& \method{} (ours)  & \textbf{98.6} & \textbf{89.2} & \textbf{98.2} & \textbf{88.8} & \textbf{97.0} & \textbf{75.7} & \textbf{96.9} & \textbf{80.5} & \textbf{97.7} & \textbf{83.5} \\
\midrule
\multirow{8}{*}{128}
& Loss~\loss & 63.5 & 11.5 & 65.3 & 14.4 & 70.3 & 17.3 & 70.0 & 22.1 & 67.3 & 16.3 \\
& Ref~\reference & 63.5 & 13.5 & 65.3 & 15.4 & 70.5 & 18.3 & 70.9 & 22.1 & 67.6 & 17.3 \\
& Zlib~\zlib & 65.3 & 16.3 & 67.2 & 19.2 & 71.5 & 19.2 & 71.2 & 18.3 & 68.8 & 18.3 \\
& Neighbor~\neighbor & 64.8 & 15.8 & 67.5 & 10.8 & 71.6 & 15.8 & 72.2 & 15.1 & 69.0 & 14.4 \\
& Min-K\%~\Mink & 66.9 & 8.7 & 69.6 & 16.3 & 76.0 & 25.0 & 73.4 & 23.1 & 71.5 & 18.3 \\
& Min-K\%++~\MinkPlusPlus & 67.1 & 9.6 & 69.2 & 17.3 & 75.2 & 20.2 & 83.4 & 21.2 & 73.7 & 17.1 \\
& ReCall~\recallcite & 88.2 & 42.3 & 90.7 & 55.8 & 90.0 & 51.9 & 91.2 & 43.3 & 90.0 & 48.3 \\
\rowcolor{gray!15}& \method{} (ours)   & \textbf{94.8} & \textbf{77.9} & \textbf{96.6} & \textbf{84.6} & \textbf{95.3} & \textbf{67.3} & \textbf{96.1} & \textbf{74.0} & \textbf{95.7} & \textbf{75.9} \\
\bottomrule
\end{tabular}
}
\caption{\textbf{AUC and TPR@5\%FPR results on WikiMIA benchmark.} \textbf{Bolded} number shows the best result within each column for the given length. \method{} achieves significant improvements over all existing baseline methods in all settings.}
\label{tab:wikimia_main_results}
\end{table*}

\paragraph{Metrics.}
Following the standard evaluation metrics~\citep{mink, minkpp, recall_mia}, we report the AUC (area under the ROC curve) to measure the trade-off between the True Positive Rate (TPR) and False Positive Rate (FPR). We also include TPR at low FPRs (TPR@5\%FPR) as an additional metrics. 

\paragraph{Implementation Details.}
For Min-K\% and Min-K\%++, we vary the hyperparameter $k$ from 10 to 100 in steps of 10. For \method{}, we optimize $\gamma$ from 0.1 to 1.0 in steps of 0.1. Following~\citet{recall_mia}, we use seven shots for both ReCall and \method{} on WikiMIA. For MIMIR, due to its increased difficulty, we vary the number of shots from 1 to 10. In all cases, we report the best performance. For more details, see Appendix~\ref{app:more_details}.

\subsection{Results}

\paragraph{Results on WikiMIA.}

Table~\ref{tab:wikimia_main_results} summarizes the experimental results on WikiMIA, demonstrating \method{}'s significant improvements over baseline methods. In terms of AUC performance, our method improved upon ReCall by 7.4\%, 6.6\%, and 5.7\% on WikiMIA-32, -64, and -128 respectively, achieving an average improvement of 6.6\% and state-of-the-art performance. For TPR@5\%FPR, \method{} outperformed the runner-up by even larger margins: 30.0\%, 34.8\%, and 27.6\% on WikiMIA-32, -64, and -128 respectively, with an average improvement of 30.8\%. Notably, \method{} achieves the best performance across models of different sizes, from Mamba-1.4B to LLaMA-30B, demonstrating its robustness and effectiveness. The consistent performance across varying sequence lengths suggests that \method{} effectively identifies membership information in both short and long text samples, underlining its potential as a powerful tool for detecting pre-training data in large language models in diverse scenarios.

\paragraph{Results on MIMIR.}

We summarize the experimental results on MIMIR in Appendix~\ref{app:mimir}.
The performance of \method{} on the MIMIR benchmark demonstrates its competitive edge across various datasets and model sizes. In the 7-gram setting, \method{} consistently achieved top-tier results, often outperforming baseline methods. Notably, on several datasets, our method frequently secured the highest scores in both AUC and TPR metrics. 
In the 13-gram setting, \method{} maintained its strong performance, particularly with larger model sizes. While overall performance decreased compared to the 7-gram setting, \emph{} still held leading positions across multiple datasets. It's worth noting that \method{} exhibited superior performance when dealing with larger models, indicating good scalability for more complex and larger language models. Although other methods occasionally showed slight advantages in certain datasets, \method's overall robust performance underscores its potential as an effective method for detecting pre-training data in large language models.

\begin{figure}[t]
\centering
\includegraphics[width=0.43\textwidth]{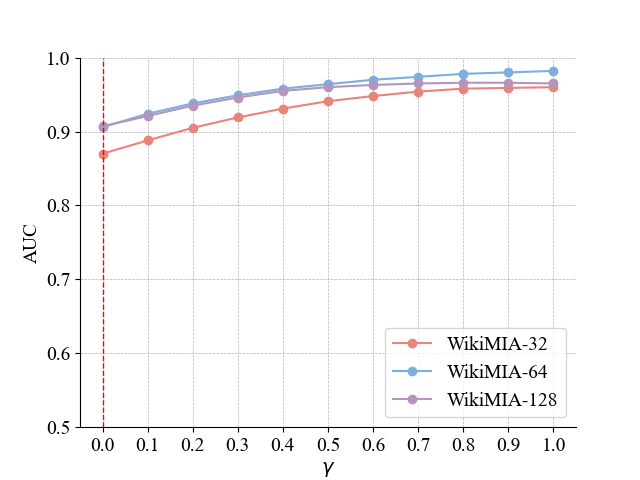}
\caption{\textbf{Ablation on $\gamma$.} 
The plot illustrates the AUC performance across different $\gamma$ values for the WikiMIA dataset. 
The red vertical line marks the $\gamma = 0$ case, where the \method{} reverts to the baseline ReCall method. As seen in this figure, \method{} ($\gamma > 0$) consistently outperforms ReCall ($\gamma = 0$).}
\label{fig:abl_gamma}
\vspace{-1em}
\end{figure}

\begin{table*}[t!]
\centering
\resizebox{1\textwidth}{!}{%
\begin{tabular}{llcccccccccccccccc}
\toprule
\multirow{4}{*}{\textbf{Len.}} & \multirow{4}{*}{\textbf{Method}} & \multicolumn{8}{c}{\textbf{Pythia-6.9B}} & \multicolumn{8}{c}{\textbf{LLaMA-30B}} \\
\cmidrule(lr){3-10} \cmidrule(lr){11-18}
 &  & \multirow{2}{*}{\textbf{Orig.}} & \multicolumn{3}{c}{\textbf{Random Del.}} & \multicolumn{3}{c}{\textbf{Synonym Sub.}} & \multirow{2}{*}{\textbf{Para.}} & \multirow{2}{*}{\textbf{Orig.}} & \multicolumn{3}{c}{\textbf{Random Del.}} & \multicolumn{3}{c}{\textbf{Synonym Sub.}} & \multirow{2}{*}{\textbf{Para.}} \\
\cmidrule(lr){4-6} \cmidrule(lr){7-9} \cmidrule(lr){12-14} \cmidrule(lr){15-17}
 &  &  & \textbf{10\%} & \textbf{15\%} & \textbf{20\%} & \textbf{10\%} & \textbf{15\%} & \textbf{20\%} &  &  & \textbf{10\%} & \textbf{15\%} & \textbf{20\%} & \textbf{10\%} & \textbf{15\%} & \textbf{20\%} &  \\
\midrule
\multirow{8}{*}{32}
& Loss~\loss & 63.7 & 60.4 & 59.6 & 56.6 & 61.5 & 59.6 & 59.5 & 63.8 & 69.4 & 66.3 & 67.0 & 64.5 & 68.4 & 66.8 & 65.8 & 70.1 \\
& Ref~\reference & 63.9 & 60.6 & 59.7 & 56.6 & 61.6 & 59.7 & 59.6 & 63.9 & 69.9 & 66.4 & 67.2 & 64.7 & 68.6 & 66.8 & 66.1 &  70.5\\
& Zlib~\zlib  & 64.4 & 61.2 & 60.2 & 58.4 & 62.2 & 60.7 & 60.8 & 64.0 & 69.9 & 66.8 & 66.9 & 64.9 & 68.8 & 67.2 & 66.5 & 70.3 \\
& 
Min-K\%~\Mink & 66.1 & 60.5 & 59.6 & 56.6 & 61.7 & 59.9 & 59.6 & 64.8 & 70.1 & 66.3 & 67.0 & 64.6 & 68.4 & 66.8 & 65.8 & 70.4 \\& 
Min-K\%++~\MinkPlusPlus & 70.0 & 59.0 & 54.5 & 51.6 & 62.5 & 59.8 & 60.1 & 67.6 & 84.6 & 71.6 & 68.2 & 67.1 & 76.9 & 73.5 & 70.1 & 81.2 \\
& ReCall~\recallcite & 87.0 & 86.2 & 83.3 & 75.2 & 88.5 & 87.5 & 83.1 & 87.8 & 91.4 & 88.1 & 88.3 & 82.7 & 87.1 & 86.4 & 84.2 & 91.0 \\
\rowcolor{gray!15}& \method{} (ours)  & \textbf{96.0} & \textbf{92.2} & \textbf{94.4} & \textbf{90.4} & \textbf{96.5} & \textbf{94.0} & \textbf{90.0} & \textbf{97.1} & \textbf{97.4} & \textbf{97.4} & \textbf{95.5} & \textbf{94.3} & \textbf{97.6} & \textbf{95.5} & \textbf{90.0} & \textbf{97.1} \\
\midrule
\multirow{8}{*}{64}
& Loss~\loss & 60.3 & 58.3 & 56.4 & 57.7 & 59.6 & 58.1 & 56.5 &  58.5 & 66.1 & 65.4 & 61.9 & 63.4 & 65.3 & 63.5 & 62.3 & 65.1 \\
& Ref~\reference & 60.4 & 58.4 & 56.5 & 57.8 & 59.6 & 58.2 & 56.6 & 58.7 & 67.0 & 65.9 & 62.2 & 63.7 & 65.9 & 64.0 & 62.7 &  65.8\\
& Zlib~\zlib & 61.6 & 60.9 & 57.8 & 60.0 & 61.8 & 59.9 & 58.2 & 60.5 & 67.1 & 67.2 & 62.6 & 65.4 & 67.1 & 65.0 & 63.5 & 66.7 \\
& Min-K\%~\Mink & 64.6 & 59.2 & 57.4 & 57.7 & 61.4 & 58.5 & 57.0 & 60.0 & 68.5 & 66.1 & 62.4 & 63.4 & 65.4 & 63.6 & 62.3 & 65.2 \\
& Min-K\%++~\MinkPlusPlus & 71.4 & 55.9 & 55.8 & 52.3 & 62.8 &  56.3 & 59.1 & 64.4 & 85.3 & 69.1 & 70.4 & 68.7 & 72.1 & 67.1 & 68.0 & 75.1 \\
& ReCall~\recallcite & 90.6 & 87.5 & 84.6 & 84.4 & 89.2 & 85.4 & 87.5 & 89.7 & 92.7 & 89.3 & 87.5 & 86.7 & 91.2 & 86.5 & 83.8 &  94.7 \\
\rowcolor{gray!15}& \method{} (ours)  & \textbf{98.2} & \textbf{96.3} & \textbf{94.3} & \textbf{96.3} & \textbf{97.7} & \textbf{95.4} & \textbf{96.6}  & \textbf{97.9} & \textbf{96.9} & \textbf{96.1} & \textbf{97.4} & \textbf{96.4} & \textbf{97.8} & \textbf{97.1} & \textbf{95.8} & \textbf{97.6} \\
\midrule
\multirow{8}{*}{128}
& Loss~\loss & 65.3 & 64.6 & 60.4 & 58.8 & 63.1 & 62.4 & 66.4 & 65.0 & 70.0 & 71.1 & 65.9 & 67.3 & 68.5 & 67.1 & 71.4 & 69.2 \\
& Ref~\reference & 65.3 & 64.8 & 60.5 & 58.9 & 63.2 & 62.4 & 66.4 & 65.0 & 70.9 & 71.6 & 66.1 & 67.5 & 69.3 & 67.4 & 72.0 & 69.8 \\
& Zlib~\zlib & 67.2 & 65.9 & 61.3 & 62.0 & 65.1 & 64.8 & 67.8 & 66.9 & 71.2 & 71.4 & 67.5 & 68.5 & 70.7 & 69.0 & 72.6 & 71.0 \\
& Min-K\%~\Mink & 69.6 & 65.2 & 60.8 & 58.8 & 65.4 & 63.0 & 66.9 & 67.4 & 73.4 & 73.2 & 67.6 & 67.9 & 70.5 & 67.3 & 72.2 & 70.4 \\
& Min-K\%++~\MinkPlusPlus & 69.2 & 55.2 & 43.0 & 45.2 & 64.8 &  49.5 & 48.6 & 65.1 & 83.4 & 68.8 & 65.5 & 64.4 & 72.0 & 62.3 &  68.3 & 73.8 \\
& ReCall~\recallcite & 90.7 & 81.8 & 80.4 & 80.0 & 89.6 & 85.5 & 84.2 & 90.4 & 91.2 & 83.2 & 78.2 & 87.3 & 81.4 & 82.4 & 82.1 & 90.9 \\
\rowcolor{gray!15}& \method{} (ours)  & \textbf{96.6} & \textbf{94.8} & \textbf{93.8} & \textbf{93.8} & \textbf{97.4} & \textbf{93.6} & \textbf{96.3}  & \textbf{95.9} & \textbf{96.1} & \textbf{95.3} & \textbf{91.1} & \textbf{99.0} & \textbf{95.6} & \textbf{92.5} & \textbf{94.2} & \textbf{95.2} \\
\bottomrule
\end{tabular}
}
\caption{\textbf{AUC performance on the WikiMIA benchmark under various text manipulation techniques.} \textbf{Bolded} numbers indicate the best result within each column for the given text length. "Orig." denotes original text without manipulation, "Random Del." refers to random deletion, "Synonym Sub." to synonym substitution, and "Para." to paraphrasing. Our method demonstrates robustness against these manipulations, consistently outperforming other baselines across different text modifications.}
\label{tab:robustness}
\end{table*}

\subsection{Ablation Study}

We focus on WikiMIA with the Pythia-6.9B model for ablation study.

\paragraph{Ablation on \(\gamma\).}

In \method{}, we introduce a hyperparameter \(\gamma\), which controls the contrastive strength between member and non-member prefixes. The AUC performance across different \(\gamma\) values for the WikiMIA dataset is depicted in Figure~\ref{fig:abl_gamma}. The red vertical lines mark the \(\gamma = 0\) case, where \method{} reverts to the baseline ReCall method.

The performance of \method{} fluctuates as $\gamma$ varies, meaning that there exist an optimal value for $\gamma$ for us to get the best performance. However, even without any fine-tuning on $\gamma$, our method still outperforms ReCall and other baselines.

\begin{figure*}
    \centering
    \begin{subfigure}{0.32\textwidth}
        \includegraphics[width=\linewidth]{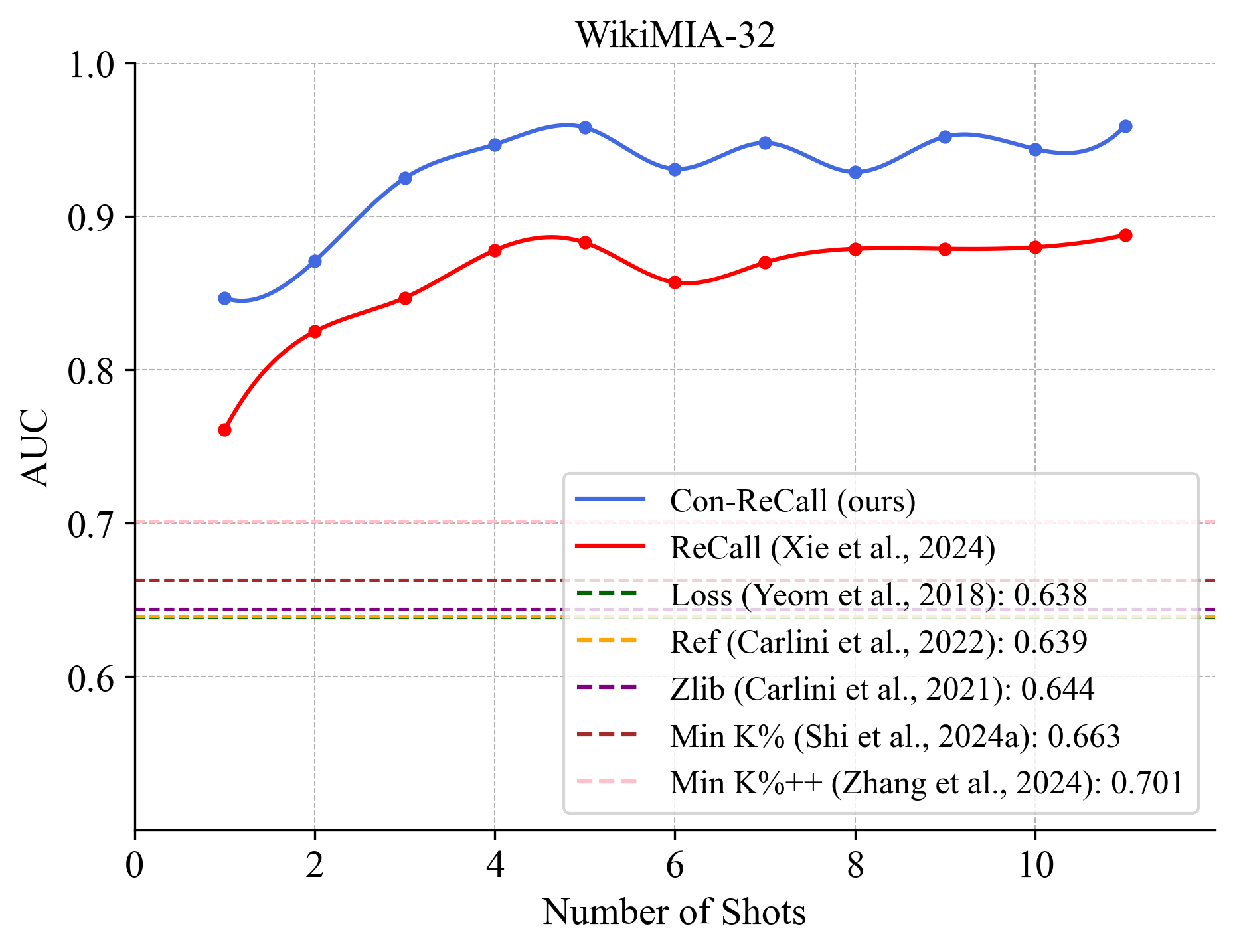}
    \end{subfigure}
    \hfill
    \begin{subfigure}{0.32\textwidth}
        \includegraphics[width=\linewidth]{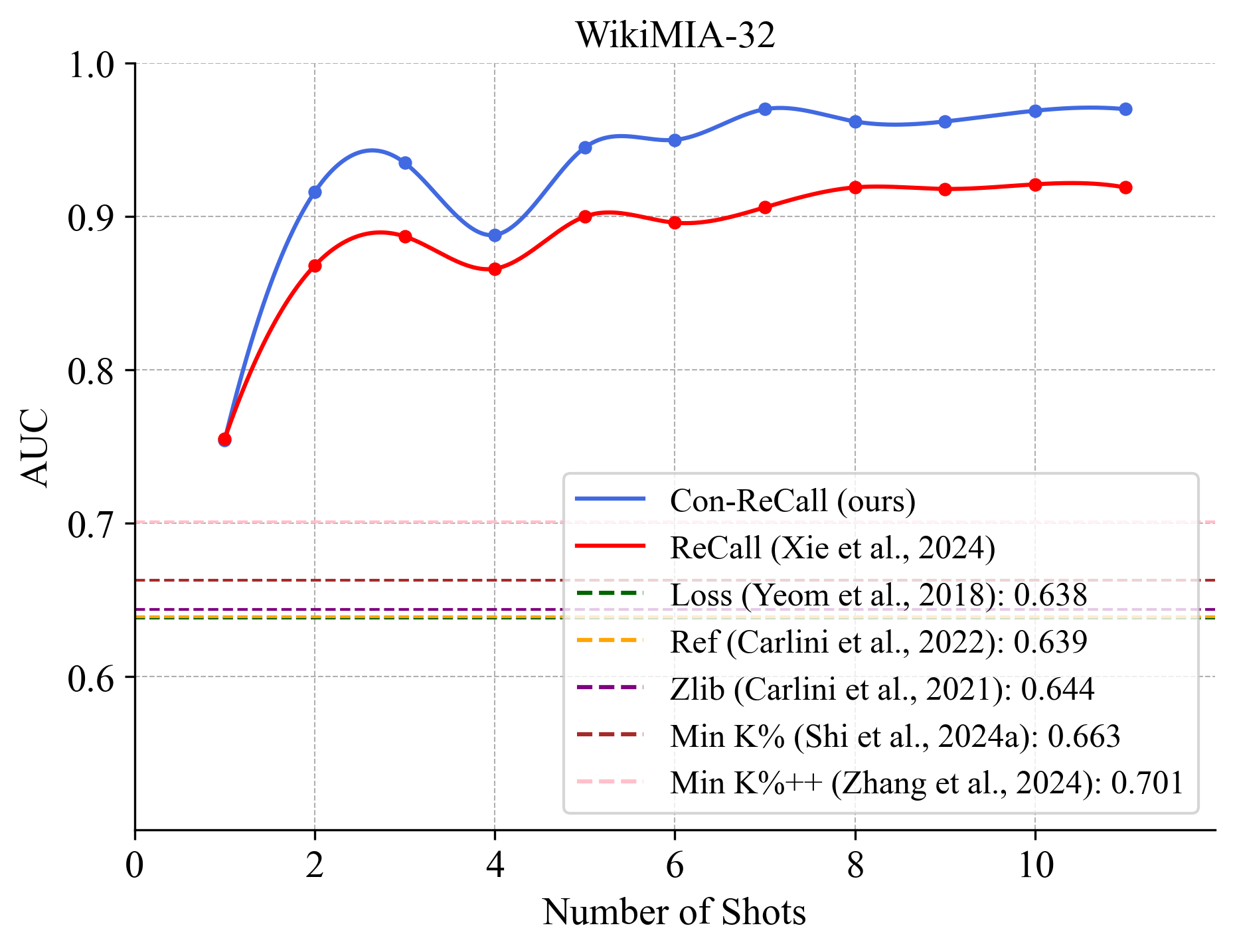}
    \end{subfigure}
    \hfill
    \begin{subfigure}{0.32\textwidth}
        \includegraphics[width=\linewidth]{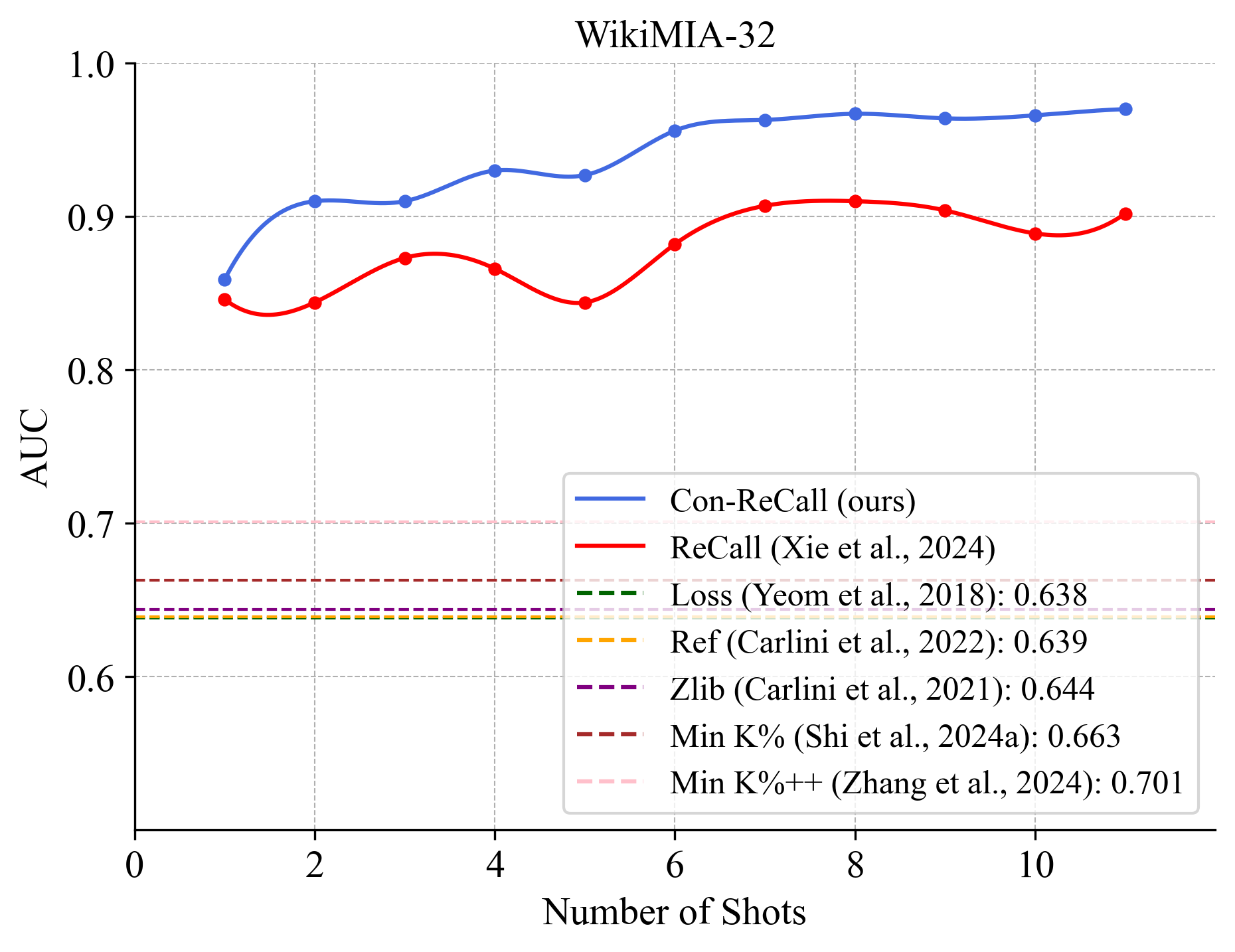}
    \end{subfigure}
    \caption{\textbf{Ablation on the number of shots.} \method{} consistently outperforms all baseline methods by a great margin on WikiMIA dataset.}
    \label{fig:abl_on_shots}
\end{figure*}

\paragraph{Ablation on the number of shots.}

The prefix is derived by concatenating a series of member or non-member strings, i.e., $P = p_{1} \oplus p_{2} \oplus \cdots \oplus p_{n}$, and we refer to the number of strings as shots following \citet{recall_mia}'s convention. In this section, we evaluate the relationship between AUC performance and the number of shots. We vary the number of shots on the WikiMIA dataset using the Pythia-6.9B model, and summarize the results in Figure~\ref{fig:abl_on_shots}. 

The general trend shows that increasing the number of shots improves the AUC, as more shots provide more information. Both ReCall and \method{} exhibit this trend, but \method{} significantly enhances the AUC compared to ReCall and outperforms all baseline methods.

\section{Analysis}
\label{sec:discussion}

To further evaluate the effectiveness and practicality of \method{}, we conducted additional analyses focusing on its robustness and adaptability in real-world scenarios.

\subsection{Robustness of \method{}}

As membership inference attacks gain prominence, evaluating their robustness against potential evasion techniques becomes crucial. Real-world data may be altered due to preprocessing, language variations, or intentional obfuscation. Therefore, a robust membership inference method should remain effective when faced with modified target data.
To assess \method{}'s robustness, we employ three text manipulation techniques. First, Random Deletion, where we randomly remove 10\%, 15\%, and 20\% of words from the original text. Second, Synonym Substitution, replacing 10\%, 15\%, and 20\% of words with their synonyms using WordNet~\citep{wordnet}. Lastly, we leverage the WikiMIA-paraphrased dataset~\citep{minkpp}, which offers ChatGPT-generated rephrased versions of the original WikiMIA~\cite{mink} texts while preserving their meaning.

We evaluate the effectiveness of baselines and \method{} after transforming texts using the above techniques. Our experiments are conducted using Pythia-6.9B~\citep{biderman2023pythiasuiteanalyzinglarge} and LLaMA-30B~\citep{touvron2023llamaopenefficientfoundation} models on the WikiMIA-32 \citep{mink} dataset. Table~\ref{tab:robustness} presents the AUC performance for each method under various text manipulation scenarios. The results demonstrate that \method{} consistently outperforms baseline methods across all text manipulation techniques, maintaining its superior performance even when faced with altered versions of the target data. This robustness underscores \method{}'s effectiveness in real-world scenarios where data may undergo various transformations.

\subsection{Approximation of Members}

In real-world scenarios, access to member data may be limited or even impossible. Therefore, it is crucial to develop methods that can approximate member data effectively. Our approach to approximating members is driven by two primary motivations. First, large language models (LLMs) are likely to retain information about significant events that occurred before their knowledge cutoff date. This retention suggests that LLMs have the potential to recall and replicate crucial aspects of such events when prompted. Second, when presented with incomplete information and tasked with its completion, LLMs can effectively leverage their internalized knowledge to generate contextually appropriate continuations. These two motivations underpin our method, where we first utilize an external LLM to enumerate major historical events. We then truncate these events and prompt the target LLM to complete them, hypothesizing that the generated content can serve as an effective approximation of the original data within the training set.

To test this approach, we first employed GPT-4o~\citep{GPT-4o} to generate descriptions of seven major events that occurred before 2020 (the knowledge cutoff date for the Pythia models). We then truncated these descriptions and prompted the target model to complete them. This method allows us to simulate the generation of data resembling the original members without directly accessing the original training set. Details of the prompts and the corresponding responses can be found in Appendix~\ref{app:approximation}.

We evaluated this method using a fixed number of seven shots for consistency with our previous experiments. The results, summarized in Table~\ref{tab:generated_prefix}, demonstrate that even without prior knowledge of actual member data, this approximation approach yields competitive results, outperforming several baseline methods.

This finding suggests that when direct access to member data is not feasible, leveraging the model's own knowledge to generate member-like content can be an effective alternative.

\begin{table}[ht!]
\resizebox{0.50\textwidth}{!}{%
\centering
\begin{tabular}{lccc}
\toprule
\textbf{Method} & \textbf{WikiMIA-32} & \textbf{WikiMIA-64} & \textbf{WikiMIA-128} \\
\midrule
Loss~\loss & 63.7 & 60.3 & 65.3 \\
Ref~\reference & 63.9 & 60.4 & 65.3 \\
Zlib~\zlib & 64.4 & 61.6 & 67.2 \\
Neighbor~\neighbor & 65.8 & 63.2 & 67.5 \\
Min-K\%~\Mink & 66.1 & 64.6 & 69.6 \\
Min-K\%++~\MinkPlusPlus  & 70.0 & 71.4 & 69.2 \\
ReCall~\recallcite & 87.0 & 90.6 & 90.7 \\
\rowcolor{gray!15} \method{} (zero access) & \underline{87.5} & \underline{91.8} & \underline{91.2} \\
\rowcolor{gray!15} \method{} (partial access) & \textbf{96.1} & \textbf{98.2} & \textbf{96.6} \\

\bottomrule
\end{tabular}
}
\caption{AUC results on WikiMIA benchmark. \textcolor{gray!100}{Gray} rows are our method and \textbf{bolded} numbers are the best performance within a column with \underline{underline} indicating the runner-up.}
\
\label{tab:generated_prefix}
\vspace{-1em}
\end{table}

\section{Conclusion}

In this paper, we introduced \method{}, a novel contrastive decoding approach for detecting pre-training data in large language models. By leveraging both member and non-member contexts, CON-RECALL significantly enhances the distinction between member and non-member data. Through extensive experiments on multiple benchmarks, we demonstrated that CON-RECALL achieves substantial improvements over existing baselines, highlighting its effectiveness in detecting pre-training data. Moreover, CON-RECALL showed robustness against various text manipulation techniques, including random deletion, synonym substitution, and paraphrasing, maintaining superior performance and resilience to potential evasion strategies. These results underscore CON-ReCall's potential as a powerful tool for addressing privacy and security concerns in large language models, while also opening new avenues for future research in this critical area.

\subsection*{Limitations}

The efficacy of \method{} is predicated on gray-box access to the language model, permitting its application to open-source models and those providing token probabilities. However, this prerequisite constrains its utility in black-box scenarios, such as API calls or online chat interfaces. Furthermore, the performance of \method{} is contingent upon the selection of member and non-member prefixes. The development of robust, automated strategies for optimal prefix selection remains an open research question. While our experiments demonstrate a degree of resilience against basic text manipulations, the method's robustness in the face of more sophisticated adversarial evasion techniques warrants further rigorous investigation.

\subsection*{Ethical Considerations}

The primary objective in developing \method{} is to address privacy and security concerns by advancing detection techniques for pre-training data in large language models. However, it is imperative to acknowledge the potential for misuse by malicious actors who might exploit this technology to reveal sensitive information. Consequently, the deployment of \method{} necessitates meticulous consideration of ethical implications and the establishment of stringent safeguards. Future work should focus on developing guidelines for the responsible use of such techniques, balancing the benefits of enhanced model transparency with the imperative of protecting individual privacy and data security.

\section*{Acknowledgement}
The work is supported by a National Science Foundation CAREER award \#2339766, a research award NSF \#2331966, University of California, Merced, University of Queensland, and the Ministry of Education, Singapore, under the Academic Research Fund Tier 1 (FY2023) (Grant A-8001996-00-00).
The views and conclusions are those of the authors and should not reflect the official policy or position of the U.S. Government.

\normalem
\bibliography{custom}

\onecolumn
\appendix

\section{Datasets Statistics}
\label{app:stats}

\begin{table}[ht]
\centering
\resizebox{0.4\textwidth}{!}{%
\begin{tabular}{lccc}
\toprule
\multirow{2}{*}{\textbf{Dataset}} & \multicolumn{3}{c}{\textbf{Text Length}} \\
\cmidrule(l){2-4}
& \textbf{32} & \textbf{64} & \textbf{128} \\
\midrule
Total Samples & 776 & 542 & 250 \\
Non-member Ratio & 50.1\% & 47.6\% & 44.4\% \\
Member Ratio & 49.9\% & 52.4\% & 55.6\% \\
\bottomrule
\end{tabular}
}
\caption{\textbf{WikiMIA Dataset Statistics.} Showing total samples and ratios for different text lengths.}
\label{tab:wikimia}
\vspace{-1em}
\end{table}

\begin{table}[ht]
\centering
\resizebox{0.4\textwidth}{!}{%
\begin{tabular}{lcc}
\toprule
\textbf{Subset} & \textbf{ngram\_7\_0.2} & \textbf{ngram\_13\_0.8} \\
\midrule
wikipedia\_(en) & 2000 & 2000 \\
github & 536 & 2000 \\
pile\_cc & 2000 & 2000 \\
pubmed\_central & 982 & 2000 \\
arxiv & 1000 & 2000 \\
dm\_mathematics & 178 & 2000 \\
hackernews & 1292 & 2000 \\
\bottomrule
\end{tabular}
}
\caption{\textbf{MIMIR Dataset Statistics.} Showing total samples for each subset and split method. All subsets have an equal 50\% split between members and non-members.}
\label{tab:mimir}
\vspace{-1em}
\end{table}

\section{Additional Implementation Details}
\label{app:more_details}

All models are obtained from Huggingface\footnote{\url{https://huggingface.co/}} and deployed with 4 NVIDIA RTX 3090 GPUs.

In our evaluation process, we carefully handled the data to ensure fair comparison across all methods. The specifics of our data handling varied between the WikiMIA and MIMIR datasets:

For the WikiMIA dataset, we selected 7 samples each from the member and non-member sets to use as prefixes. The number of shots was fixed at 7 for all experiments on this dataset. 

For the MIMIR dataset, we removed 10 samples each from the member and non-member datasets to create our prefix pool. Unlike WikiMIA, we varied the number of shots from 1 to 10 and reported the best-performing configuration.

For both datasets, the samples used for prefixes were removed from the evaluation set for all methods, including baselines, ensuring a fair comparison across different methods.

\section{Member Approximation Details}
\label{app:approximation}

In this section, we detail our method for approximating member data when direct access to the original training set is not feasible. Our approach involves two steps: first, using GPT-4o~\citep{GPT-4o} to generate descriptions of significant events, and then using these partially truncated descriptions to prompt our target model.

We begin by providing GPT-4o with the following prompt:

\begin{tcolorbox}[title=Prompt for GPT-4o]
Provide a concise overview of seven pivotal global events that occurred before 2020. Each event should be described in under 32 words and presented in a narrative format without explicit headings.
\end{tcolorbox}

GPT-4o generated the following response:

\begin{tcolorbox}[title=Response from GPT-4o]
The world witnessed the fall of the Berlin Wall in 1989, symbolizing the end of the Cold War. The 9/11 attacks in 2001 reshaped global security dynamics. The 2008 financial crisis led to worldwide economic instability. In 1945, World War II concluded, altering geopolitical landscapes. The 1969 Moon landing marked humanity's leap into space. The 1979 Iranian Revolution transformed Middle Eastern politics. The 1918 Spanish flu pandemic highlighted vulnerabilities in global health.
\end{tcolorbox}

We then truncated these responses to create partial prompts:

\begin{tcolorbox}[title=Truncated Responses]
\begin{itemize}
    \item The world witnessed the fall of the Berlin Wall in 
    \item The 9/11 attacks in 
    \item The 2008 financial crisis led to 
    \item In 1945, World War II concluded, 
    \item The 1969 Moon landing marked 
    \item The 1979 Iranian Revolution transformed Middle Eastern 
    \item The 1918 Spanish flu pandemic highlighted 
\end{itemize}
\end{tcolorbox}

These truncated texts were then used as prompts for our target model to complete, simulating the generation of member-like content. To ensure consistency with our experimental setup, we set the maximum number of new tokens (max\_new\_tokens) to match the length of the target text. For example, when working with WikiMIA-32, max\_new\_tokens was set to 32.

\section{MIMIR Results}
\label{app:mimir}
\begin{table*}[ht!]
\begin{center} \scriptsize
\label{app:mimir_main_results_7}
\vspace{3em}
\subsection{MIRMIR 7-gram Results} 
\vspace{1em}
\setlength{\tabcolsep}{0.7pt}
\begin{tabularx}{\textwidth}{l *{24}{>{\centering\arraybackslash}X}@{}}
    \toprule
    \multirow{2}{*}{}  & \multicolumn{6}{c}{\textbf{Wikipedia}} & \multicolumn{6}{c}{\textbf{Github}} & \multicolumn{6}{c}{\textbf{Pile CC}} & \multicolumn{6}{c}{\textbf{PubMed Central}} \\
    \cmidrule(lr){2-7}  \cmidrule(lr){8-13} \cmidrule(lr){14-19} \cmidrule(lr){20-25}
    \textbf{Method} & \multicolumn{2}{c}{2.8B} & \multicolumn{2}{c}{6.9B} & \multicolumn{2}{c}{12B}
    & \multicolumn{2}{c}{2.8B} & \multicolumn{2}{c}{6.9B} & \multicolumn{2}{c}{12B}
    & \multicolumn{2}{c}{2.8B} & \multicolumn{2}{c}{6.9B} & \multicolumn{2}{c}{12B}
    & \multicolumn{2}{c}{2.8B} & \multicolumn{2}{c}{6.9B} & \multicolumn{2}{c}{12B}
    \\
    \cmidrule(lr){2-3} \cmidrule(lr){4-5} \cmidrule(lr){6-7}
    \cmidrule(lr){8-9} \cmidrule(lr){10-11} \cmidrule(lr){12-13}
    \cmidrule(lr){14-15} \cmidrule(lr){16-17} \cmidrule(lr){18-19}
    \cmidrule(lr){20-21} \cmidrule(lr){22-23} \cmidrule(lr){24-25}
    &  AUC &  TPR
    &  AUC &  TPR
    &  AUC &  TPR
    &  AUC &  TPR
    &  AUC &  TPR
    &  AUC &  TPR
    &  AUC &  TPR
    &  AUC &  TPR
    &  AUC &  TPR
    &  AUC &  TPR
    &  AUC &  TPR
    &  AUC &  TPR
    \\
    \midrule
    
Loss & 66.4 & 22.9 & 68.0 & 24.1 & 69.1 & 24.2 & 88.1 & 56.6 & 89.0 & 61.6 & 89.5 & 62.8 & \underline{54.8} & 11.2 & 55.9 & 13.6 & 56.4 & 13.9 & 77.9 & 31.2 & 78.0 & 31.0 & 77.9 & 32.2 \\
Ref & \underline{66.5} & \textbf{23.2} & 68.2 & \underline{23.9} & 69.2 & 24.0 & \underline{88.4} & \underline{60.9} & 89.4 & 66.3 & 89.8 & 69.0 & \underline{54.8} & \underline{11.7} & 56.0 & 13.5 & 56.4 & 13.8 & 77.7 & 30.4 & 77.8 & 30.1 & 77.6 & 32.2 \\
Zlib & 63.3 & 19.9 & 65.2 & 20.8 & 66.4 & 22.4 & \textbf{90.7} & \textbf{71.7} & \textbf{91.4} & \textbf{74.0} & \textbf{91.8} & \textbf{75.6} & 53.6 & \textbf{12.1} & 54.7 & 13.6 & 55.0 & 14.2 & 76.9 & 29.9 & 77.1 & 28.7 & 77.0 & 29.7 \\
Min-K\% & \textbf{66.6} & \underline{22.3} & \underline{68.4} & \textbf{24.4} & \underline{69.7} & 25.5 & 88.1 & 55.4 & 89.1 & 59.7 & 89.7 & 63.2 & \textbf{54.9} & 10.6 & \underline{56.3} & 12.1 & 56.5 & 13.8 & 78.6 & 33.9 & 79.0 & 33.1 & \underline{79.0} & 33.7 \\
Min-K\%++ & 65.7 & 21.1 & \textbf{69.2} & 23.1 & \textbf{71.1} & \textbf{26.2} & 85.7 & 54.7 & 86.2 & 55.8 & 87.6 & 58.9 & 54.5 & 10.9 & \textbf{56.5} & 11.1 & \textbf{56.9} & 12.1 & 68.4 & 20.2 & 70.1 & 25.2 & 70.4 & 22.9 \\
ReCall & 65.5 & 21.7 & 67.6 & 22.9 & 69.2 & 25.1 & 88.0 & 60.5 & 90.1 & 71.7 & 90.7 & \underline{72.1} & 53.8 & 9.3 & 55.6 & \textbf{14.5} & \underline{56.7} & \underline{14.6} & \textbf{79.8} & \textbf{42.6} & \textbf{81.8} & \textbf{46.8} & \textbf{79.2} & \textbf{38.5} \\
\rowcolor{gray!15}\method{} & 65.6 & 21.7 & 67.6 & 23.1 & 69.1 & \underline{25.6} & 88.0 & 57.8 & \underline{90.5} & \underline{72.1} & \underline{90.9} & \textbf{75.6} & 53.6 & 9.9 & 55.5 & \underline{14.2} & \textbf{56.9} & \textbf{14.8} & \underline{79.6} & \underline{41.8} & \underline{81.7} & \underline{46.6} & 78.7 & \underline{38.0} \\
    \toprule
    \multirow{2}{*}{}  & \multicolumn{6}{c}{\textbf{ArXiv}} & \multicolumn{6}{c}{\textbf{DM Mathematics}} & \multicolumn{6}{c}{\textbf{HackerNews}} & \multicolumn{6}{c}{\textbf{Average}}\\
    \cmidrule(lr){2-7}  \cmidrule(lr){8-13} \cmidrule(lr){14-19} \cmidrule(lr){20-25}
    \textbf{Method} & \multicolumn{2}{c}{2.8B} & \multicolumn{2}{c}{6.9B} & \multicolumn{2}{c}{12B}
    & \multicolumn{2}{c}{2.8B} & \multicolumn{2}{c}{6.9B} & \multicolumn{2}{c}{12B}
    & \multicolumn{2}{c}{2.8B} & \multicolumn{2}{c}{6.9B} & \multicolumn{2}{c}{12B}
    & \multicolumn{2}{c}{2.8B} & \multicolumn{2}{c}{6.9B} & \multicolumn{2}{c}{12B}
    \\
    \cmidrule(lr){2-3} \cmidrule(lr){4-5} \cmidrule(lr){6-7}
    \cmidrule(lr){8-9} \cmidrule(lr){10-11} \cmidrule(lr){12-13}
    \cmidrule(lr){14-15} \cmidrule(lr){16-17} \cmidrule(lr){18-19}
    \cmidrule(lr){20-21} \cmidrule(lr){22-23} \cmidrule(lr){24-25}
    &  AUC &  TPR
    &  AUC &  TPR
    &  AUC &  TPR
    &  AUC &  TPR
    &  AUC &  TPR
    &  AUC &  TPR
    &  AUC &  TPR
    &  AUC &  TPR
    &  AUC &  TPR
    &  AUC &  TPR
    &  AUC &  TPR
    &  AUC &  TPR
    \\
    \midrule
Loss & 78.0 & 34.1 & \underline{79.0} & \textbf{36.7} & \textbf{79.5} & \underline{36.1} & 91.3 & 58.2 & 91.4 & 58.2 & 91.3 & 59.5 & 60.6 & \underline{11.0} & \underline{61.3} & \underline{11.9} & 62.1 & \underline{14.3} & 73.9 & 32.2 & \underline{74.7} & 33.9 & 75.1 & 34.7 \\
Ref & 78.0 & 34.5 & \textbf{79.1} & \underline{36.1} & \textbf{79.5} & \textbf{36.7} & 89.8 & 41.8 & 89.9 & 43.0 & 89.7 & 41.8 & 60.6 & \underline{11.0} & \underline{61.3} & \underline{11.9} & \underline{62.2} & \textbf{14.5} & 73.7 & 30.5 & 74.5 & 32.1 & 74.9 & 33.1 \\
Zlib & 77.5 & 35.1 & 78.4 & 34.1 & \underline{78.7} & 34.7 & 80.2 & 16.5 & 80.4 & 16.5 & 80.4 & 16.5 & 59.2 & 10.4 & 59.6 & 10.4 & 60.2 & 12.3 & 71.6 & 27.9 & 72.4 & 28.3 & 72.8 & 29.3 \\
Min-K\% & 78.0 & 34.1 & \underline{79.0} & \textbf{36.7} & \textbf{79.5} & \underline{36.1} & 93.3 & 69.6 & \underline{93.2} & 68.4 & \underline{93.1} & 69.6 & \underline{60.6} & \underline{11.0} & \underline{61.3} & 11.8 & \underline{62.2} & \underline{14.3} & 74.3 & 33.8 & \textbf{75.2} & 35.2 & \textbf{75.7} & 36.6 \\
Min-K\%++ & 66.7 & 16.7 & 69.4 & 17.8 & 70.7 & 19.0 & 77.4 & 30.4 & 75.1 & 27.8 & 76.4 & 22.8 & 58.3 & 8.5 & 59.7 & 8.5 & 61.2 & 8.3 & 68.1 & 23.2 & 69.5 & 24.2 & 70.6 & 24.3 \\
ReCall & \underline{79.5} & \underline{36.5} & 77.0 & 31.8 & 78.0 & 32.2 & \underline{94.4} & \underline{87.3} & 92.9 & \textbf{81.0} & 92.2 & \underline{72.2} & 60.4 & 10.7 & \textbf{61.4} & \textbf{12.4} & \textbf{62.4} & 10.8 & \underline{74.5} & \underline{38.4} & \textbf{75.2} & \textbf{40.2} & \underline{75.5} & \underline{37.9} \\
\rowcolor{gray!15}\method{} & \textbf{80.4} & \textbf{42.0} & 77.1 & 31.2 & 78.5 & 31.2 & \textbf{95.2} & \textbf{88.6} & \textbf{93.3} & \underline{77.2} & \textbf{93.6} & \textbf{83.5} & \textbf{60.7} & \textbf{12.3} & 60.8 & 9.6 & 61.7 & 10.7 & \textbf{74.7} & \textbf{39.2} & \textbf{75.2} & \underline{39.1} & \textbf{75.7} & \textbf{39.9} \\
\bottomrule
\end{tabularx}
\caption{\textbf{AUC and TPR (TPR@5\%FPR) results on the MIMIR benchmark in the 7-gram setting.} \textbf{Bolded} numbers indicate the best result within each column, with the runner-up \underline{underlined}. Our method  demonstrates competitive performance across various datasets and model sizes, frequently achieving top or near-top results in both AUC and TPR metrics.}
\end{center}
\end{table*}
\begin{table*}[ht!]
\begin{center} \scriptsize
\label{app:mimir_main_results_13}
\vspace{3em}
\subsection{MIMIR 13-gram Results} 
\vspace{1em}
\setlength{\tabcolsep}{0.7pt}
\begin{tabularx}{\textwidth}{l *{24}{>{\centering\arraybackslash}X}@{}}
    \toprule
    \multirow{2}{*}{}  & \multicolumn{6}{c}{\textbf{Wikipedia}} & \multicolumn{6}{c}{\textbf{Github}} & \multicolumn{6}{c}{\textbf{Pile CC}} & \multicolumn{6}{c}{\textbf{PubMed Central}} \\
    \cmidrule(lr){2-7}  \cmidrule(lr){8-13} \cmidrule(lr){14-19} \cmidrule(lr){20-25}
    \textbf{Method} & \multicolumn{2}{c}{2.8B} & \multicolumn{2}{c}{6.9B} & \multicolumn{2}{c}{12B}
    & \multicolumn{2}{c}{2.8B} & \multicolumn{2}{c}{6.9B} & \multicolumn{2}{c}{12B}
    & \multicolumn{2}{c}{2.8B} & \multicolumn{2}{c}{6.9B} & \multicolumn{2}{c}{12B}
    & \multicolumn{2}{c}{2.8B} & \multicolumn{2}{c}{6.9B} & \multicolumn{2}{c}{12B}
    \\
    \cmidrule(lr){2-3} \cmidrule(lr){4-5} \cmidrule(lr){6-7}
    \cmidrule(lr){8-9} \cmidrule(lr){10-11} \cmidrule(lr){12-13}
    \cmidrule(lr){14-15} \cmidrule(lr){16-17} \cmidrule(lr){18-19}
    \cmidrule(lr){20-21} \cmidrule(lr){22-23} \cmidrule(lr){24-25}
    &  AUC &  TPR
    &  AUC &  TPR
    &  AUC &  TPR
    &  AUC &  TPR
    &  AUC &  TPR
    &  AUC &  TPR
    &  AUC &  TPR
    &  AUC &  TPR
    &  AUC &  TPR
    &  AUC &  TPR
    &  AUC &  TPR
    &  AUC &  TPR
    \\
    \midrule
    
Loss & 51.9 & 4.6 & 52.9 & 5.1 & 53.6 & 5.2 & 71.4 & 33.4 & 73.1 & 38.5 & 74.1 & 40.2 & 50.2 & 4.9 & 50.8 & 4.9 & 51.2 & 5.2 & 49.9 & 4.2 & 50.6 & 4.5 & 51.3 & 5.1 \\
Ref & 52.0 & 4.9 & 53.0 & 6.0 & 53.7 & 5.8 & 70.5 & 25.6 & 71.9 & 26.6 & 72.5 & 27.2 & 50.2 & 5.1 & 50.8 & 5.1 & 51.2 & 5.4 & 49.9 & 4.3 & 50.7 & 4.1 & 51.4 & 4.8 \\
Zlib & \underline{52.6} & \underline{6.0} & 53.6 & 6.4 & 54.4 & 6.8 & \textbf{72.4} & \textbf{36.3} & \underline{74.1} & 39.4 & \textbf{75.0} & 40.9 & 50.2 & \underline{5.5} & 50.8 & \underline{6.3} & 51.1 & \textbf{6.7} & 50.1 & 3.5 & 50.7 & 4.0 & 51.2 & 4.4 \\
Min-K\% & 51.9 & 5.2 & 53.6 & \underline{6.6} & 54.5 & \underline{8.1} & 71.5 & 33.4 & 73.3 & 37.3 & \underline{74.3} & 39.1 & 50.8 & 3.9 & 51.5 & 4.5 & 51.7 & 4.8 & 50.4 & 4.5 & 51.2 & 5.2 & 52.4 & \underline{4.9} \\
Min-K\%++ & \textbf{55.1} & \textbf{6.2} & \textbf{58.0} & \textbf{9.2} & \textbf{60.9} & \textbf{11.1} & 70.9 & 33.9 & 72.9 & 38.1 & 74.2 & 40.0 & \underline{51.2} & 4.8 & \textbf{53.3} & 5.1 & \textbf{53.8} & 5.9 & \textbf{52.8} & \textbf{6.5} & \textbf{55.1} & \textbf{6.5} & \textbf{55.7} & \textbf{8.2} \\
ReCall & 52.5 & 3.4 & 54.7 & 4.9 & 55.3 & 5.3 & 71.4 & 34.1 & \textbf{74.5} & \textbf{42.4} & \textbf{75.0} & \underline{41.9} & 50.2 & 4.3 & \underline{51.8} & 5.3 & 51.8 & \underline{6.0} & 51.5 & 4.2 & \underline{52.5} & 5.2 & \underline{53.4} & 3.9 \\
\rowcolor{gray!15} \method & 52.5 & 3.4 & \underline{54.8} & 5.3 & \underline{55.6} & 5.3 & \underline{71.7} & \underline{35.1} & \textbf{74.5} & \underline{42.3} & \textbf{75.0} & \textbf{42.1} & \textbf{52.3} & \textbf{6.4} & \textbf{53.3} & \textbf{7.5} & \underline{52.4} & 5.7 & \underline{51.8} & \underline{4.9} & \underline{52.5} & \underline{5.4} & 53.3 & 4.2 \\
    
    \toprule
    \multirow{2}{*}{}  & \multicolumn{6}{c}{\textbf{ArXiv}} & \multicolumn{6}{c}{\textbf{DM Mathematics}} & \multicolumn{6}{c}{\textbf{HackerNews}} & \multicolumn{6}{c}{\textbf{Average}}\\
    \cmidrule(lr){2-7}  \cmidrule(lr){8-13} \cmidrule(lr){14-19} \cmidrule(lr){20-25}
    \textbf{Method} & \multicolumn{2}{c}{2.8B} & \multicolumn{2}{c}{6.9B} & \multicolumn{2}{c}{12B}
    & \multicolumn{2}{c}{2.8B} & \multicolumn{2}{c}{6.9B} & \multicolumn{2}{c}{12B}
    & \multicolumn{2}{c}{2.8B} & \multicolumn{2}{c}{6.9B} & \multicolumn{2}{c}{12B}
    & \multicolumn{2}{c}{2.8B} & \multicolumn{2}{c}{6.9B} & \multicolumn{2}{c}{12B}
    \\
    \cmidrule(lr){2-3} \cmidrule(lr){4-5} \cmidrule(lr){6-7}
    \cmidrule(lr){8-9} \cmidrule(lr){10-11} \cmidrule(lr){12-13}
    \cmidrule(lr){14-15} \cmidrule(lr){16-17} \cmidrule(lr){18-19}
    \cmidrule(lr){20-21} \cmidrule(lr){22-23} \cmidrule(lr){24-25}
    &  AUC &  TPR
    &  AUC &  TPR
    &  AUC &  TPR
    &  AUC &  TPR
    &  AUC &  TPR
    &  AUC &  TPR
    &  AUC &  TPR
    &  AUC &  TPR
    &  AUC &  TPR
    &  AUC &  TPR
    &  AUC &  TPR
    &  AUC &  TPR
    \\
    \midrule
Loss & 52.0 & 4.6 & 53.0 & 5.1 & 53.5 & 5.6 & 48.5 & 4.1 & 48.6 & 4.1 & 48.6 & 3.9 & 51.1 & 5.6 & 51.9 & 6.0 & 52.6 & \underline{6.9} & 53.6 & 8.8 & 54.4 & 9.7 & 55.0 & 10.3 \\
Ref & 52.1 & 4.5 & 53.1 & 5.2 & 53.6 & 5.4 & 48.4 & 4.3 & 48.5 & 3.8 & 48.5 & 4.3 & 51.2 & 5.6 & 52.0 & 5.9 & 52.7 & 6.6 & 53.5 & 7.8 & 54.3 & 8.1 & 54.8 & 8.5 \\
Zlib & 51.4 & 4.1 & 52.3 & 4.5 & 52.7 & 4.7 & 48.1 & \underline{4.6} & 48.1 & 4.4 & 48.1 & 4.5 & 50.8 & 5.8 & 51.2 & 5.7 & 51.6 & 5.8 & 53.7 & \underline{9.4} & 54.4 & 10.1 & 54.9 & 10.5 \\
Min-K\% & 52.6 & 4.2 & 53.7 & 4.5 & 54.7 & 5.2 & 49.6 & 4.5 & 49.8 & 4.4 & 49.8 & \underline{5.2} & 52.4 & 5.9 & 53.5 & \underline{6.3} & 54.6 & 6.6 & 54.2 & 8.8 & 55.2 & 9.8 & 56.0 & 10.6 \\
Min-K\%++ & \textbf{53.8} & \underline{6.3} & \underline{55.1} & \underline{7.9} & \textbf{57.9} & 8.2 & \textbf{51.9} & \textbf{5.7} & \textbf{51.9} & \textbf{6.4} & \textbf{52.1} & \textbf{6.8} & 52.5 & 4.5 & 54.4 & \textbf{6.5} & \textbf{56.5} & 4.9 & \textbf{55.5} & \textbf{9.7} & \textbf{57.2} & \textbf{11.4} & \textbf{58.7} & \textbf{12.2} \\
ReCall & 52.9 & \textbf{6.4} & \textbf{55.7} & \textbf{8.1} & 56.6 & \textbf{8.8} & 49.5 & 3.8 & 49.9 & 3.7 & 49.5 & 3.9 & \underline{52.7} & \underline{5.9} & \textbf{54.8} & 5.5 & \underline{55.4} & \textbf{7.0} & 54.4 & 8.9 & 56.3 & 10.7 & 56.7 & \underline{11.0} \\
\rowcolor{gray!15}\method{} & \underline{52.9} & 6.0 & \textbf{55.7} & 7.5 & \underline{56.7} & \underline{8.5} & \underline{50.9} & 3.5 & \underline{50.5} & \underline{5.4} & \underline{51.5} & 4.6 & \textbf{52.8} & \textbf{6.8} & \underline{54.7} & 5.1 & \underline{55.4} & 6.6 & \underline{55.0} & \underline{9.4} & \underline{56.6} & \underline{11.2} & \underline{57.1} & \underline{11.0} \\
\bottomrule
\end{tabularx}
\caption{\textbf{AUC and TPR (TPR@5\%FPR) results on the MIMIR benchmark in the 13-gram setting.} \textbf{Bolded} numbers indicate the best result within each column, with the runner-up \underline{underlined}. Our method demonstrates strong performance across various datasets and model sizes, frequently achieving top-tier results in both AUC and TPR metrics, with particular strength in larger model sizes and specific datasets.}
\end{center}
\end{table*}
\end{document}